\documentclass[letterpaper, 10 pt, conference]{ieeeconf}  % Comment this line out if you need a4paper

\IEEEoverridecommandlockouts                              % This command is only needed if 
\usepackage{graphicx} % for pdf, bitmapped graphics files
\usepackage{color}
\usepackage{xcolor}
\usepackage{tabu}
\usepackage{amsmath}
\usepackage{rotating}
\usepackage{booktabs}

\makeatletter
\let\NAT@parse\undefined
\makeatother
\usepackage{hyperref}  %hyperref still needs to be put at the end!

\usepackage{algorithm}
\usepackage[noend]{algpseudocode}
\makeatletter
\def\BState{\State\hskip-\ALG@thistlm}
\makeatother

\definecolor{dcn}                 {RGB}{227,111, 38}
\definecolor{dcn-ft-0005}         {RGB}{128, 64,128}
\definecolor{dcn-ft-001}          {RGB}{ 62,135,207}
\definecolor{dcn-ft-002}          {RGB}{190,153,153}
\definecolor{dcn-ft-003}          {RGB}{107,142, 35}
\definecolor{dcn-ft-006}          {RGB}{200,  0,  0}

\definecolor{road}                {RGB}{128, 64,128}
\definecolor{sidewalk}            {RGB}{244, 35,232}
\definecolor{building}            {RGB}{ 70, 70, 70}
\definecolor{wall}                {RGB}{102,102,156}
\definecolor{fence}               {RGB}{190,153,153}
\definecolor{pole}                {RGB}{153,153,153}
\definecolor{traffic light}       {RGB}{250,170, 30}
\definecolor{traffic sign}        {RGB}{220,220,  0}
\definecolor{vegetation}          {RGB}{107,142, 35}
\definecolor{terrain}             {RGB}{152,251,152}
\definecolor{sky}                 {RGB}{ 70,130,180}
\definecolor{person}              {RGB}{220, 20, 60}
\definecolor{rider}               {RGB}{255,  0,  0}
\definecolor{car}                 {RGB}{  0,  0,142}
\definecolor{truck}               {RGB}{  0,  0, 70}
\definecolor{bus}                 {RGB}{  0, 60,100}
\definecolor{train}               {RGB}{  0, 80,100}
\definecolor{motorcycle}          {RGB}{  0,  0,230}
\definecolor{bicycle}             {RGB}{119, 11, 32}
\definecolor{void}                {RGB}{  0,  0,  0}

\definecolor{fine}{RGB}{62,135,207}
\definecolor{coarse}{RGB}{220,20,60}

\makeatletter
\DeclareRobustCommand\onedot{\futurelet\@let@token\@onedot}
\def\@onedot{\ifx\@let@token.\else.\null\fi\xspace}

\makeatother
\usepackage{tikz}
\usepackage{pgfplots}
\usetikzlibrary{backgrounds}

\title{\LARGE \bf
Dark Model Adaptation: Semantic Image Segmentation from Daytime to Nighttime
}

\author{Dengxin Dai$^{1}$ and Luc Van Gool$^{1,2}$% <-this % stops a space
\thanks{$^{1}$Dengxin Dai and Luc Van Gool are with the Toyota TRACE-Zurich team at the Computer Vision Lab, 
        ETH Zurich, 8092 Zurich, Switzerland
        {\tt\small firstname.lastname@vision.ee.ethz.ch }}%
\thanks{$^{2}$Luc Van Gool is also with the Toyota TRACE-Leuven team at the Dept of Electrical Engineering ESAT, KU Leuven
         3001 Leuven, Belgium}%
}

\begin{document}

\maketitle
\thispagestyle{empty}
\pagestyle{empty}

%%%%%%%%%%%%%%%%%%%%%%%%%%%%%%%%%%%%%%%%%%%%%%%%%%%%%%%%%%%%%%%%%%%%%%%%%%%%%%%%
\begin{abstract}
This work addresses the problem of semantic image segmentation of nighttime scenes. Although considerable progress has been made in semantic image segmentation, it is mainly related to daytime scenarios. This paper proposes a novel method to \emph{progressive} adapt the semantic models trained on daytime scenes, along with large-scale annotations therein, to nighttime scenes via the bridge of twilight time --- the time between dawn and sunrise, or between sunset and dusk. The goal of the method is to alleviate the cost of human annotation for nighttime images by transferring knowledge from standard daytime conditions. In addition to the method, a new dataset of road scenes is compiled; it consists of 35,000 images ranging from daytime to twilight time and to nighttime. Also, a subset of the nighttime images are densely annotated for method evaluation. Our experiments show that our method is effective for knowledge transfer from daytime scenes to nighttime scenes, without using extra human annotation. 
\end{abstract}

%%%%%%%%%%%%%%%%%%%%%%%%%%%%%%%%%%%%%%%%%%%%%%%%%%%%%%%%%%%%%%%%%%%%%%%%%%%%%%%%
\section{INTRODUCTION}
Autonomous vehicles will have a substantial impact on people's daily life, both personally and professionally. For instance, automated vehicles can largely increase human productivity by turning driving time into working time, provide personalized mobility to non-drivers, reduce traffic accidents, or free up parking space and generalize valet service~\cite{autonomous:vehicle:guide:policymakers}. As such, developing automated vehicles is becoming the core interest of many, diverse industrial players. Recent years have witnessed great progress in autonomous driving \cite{drive:surroundview:route:planner}, resulting in announcements that autonomous vehicles have driven over many thousands of miles and that companies aspire to sell such vehicles in a few years. All this has fueled expectations that fully automated vehicles are coming soon. Yet, significant technical obstacles must be overcome before assisted driving can be turned into full-fletched automated driving, a prerequisite for the above visions to materialize.

While perception algorithms based on visible light cameras are constantly getting better, they are mainly designed to operate on images taken at daytime under good illumination~\cite{vision:atmosphere,semantic:foggy:scene}. Yet, outdoor applications can hardly escape from challenging weather and illumination conditions. One of the big reasons that automated cars have not gone mainstream yet is because it cannot deal well with nighttime and adverse weather conditions. Camera sensors can become untrustworthy at nighttime, in foggy weather, and in wet weather. Thus, computer vision systems have to function well also under these adverse conditions. In this work, we focus on semantic object recognition for nighttime driving scenes.

Robust object recognition using visible light cameras remains a difficult problem. This is because the structural, textural and/or color features needed for object recognition sometimes do not exist or highly disbursed by artificial lights, to the point where it is difficult to recognize the objects even for human. The problem is further compounded by camera noise~\cite{nighttime:noise:reduction:16} and motion blur. Due to this reason, there are systems using  far-infrared (FIR) cameras instead of the widely used visible light cameras for nighttime scene understanding~\cite{night:vision:pedestrian:05,day:night:16}. 
Far-infrared (FIR) cameras can be another choice~\cite{night:vision:pedestrian:05,day:night:16}. They, however, are expensive and only provide images of relatively low-resolution. Thus, this work adopts visible light cameras for semantic segmentation of nighttime road scenes. Another reason of this choice is that large-scale datasets are available for daytime images by visible light cameras~\cite{Cityscapes}. This makes model adaptation from daytime to nighttime feasible. 

\begin{figure*}[t]
\includegraphics[trim={0 -0.03cm 0 0},clip,width=0.97\linewidth]{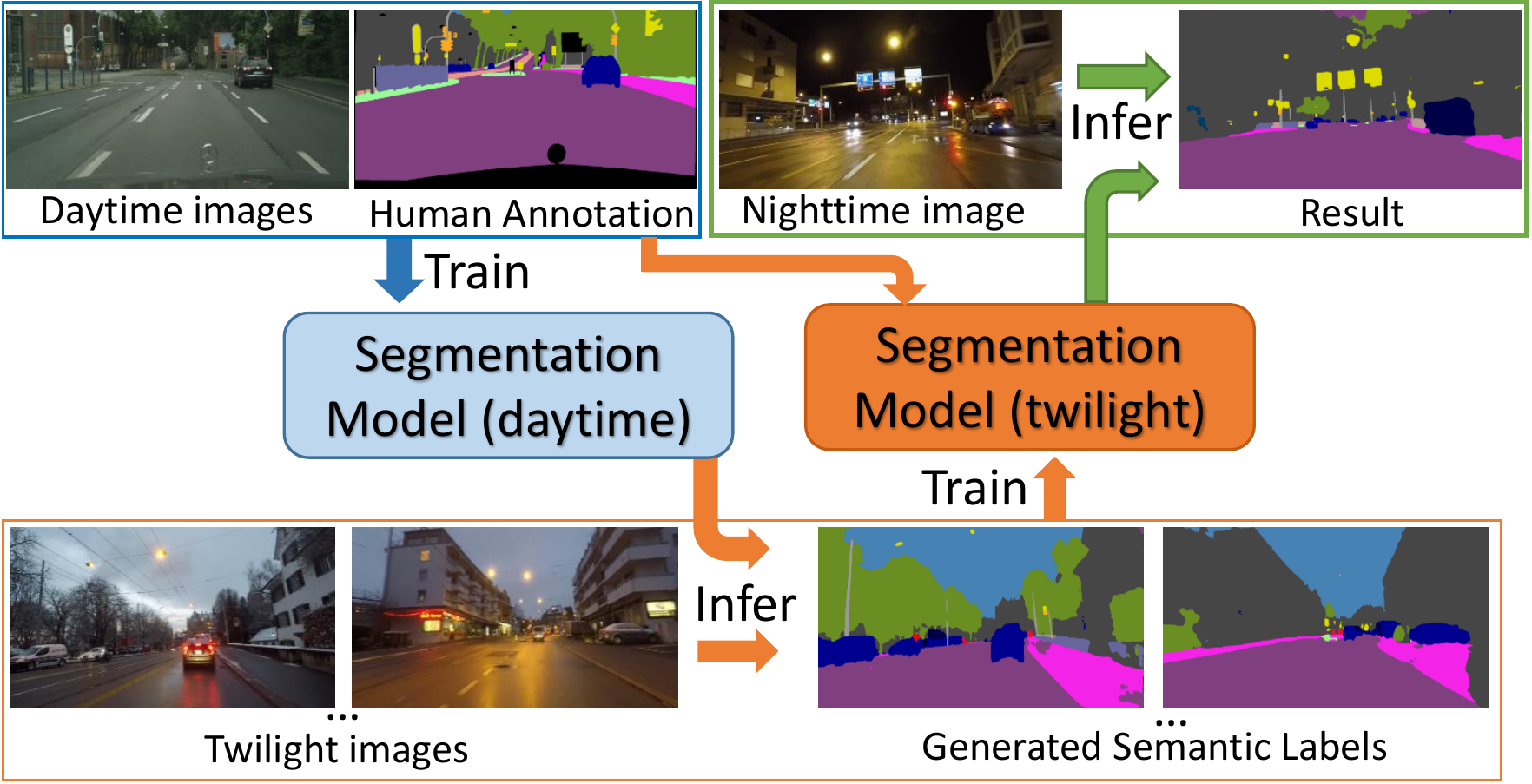} 
\resizebox{0.95\linewidth}{!}{
\begin{tikzpicture}[tight background, scale=0.75, every node/.style={font=\large}]
	\draw[white, fill=void, draw=white] (0,0) rectangle (1 * 4, 1) node[pos=0.5] {Void};
	\draw[white, fill=road, draw=white] (1 * 4,0) rectangle (2 * 4, 1) node[pos=0.5] {Road};
	\draw[white, fill=sidewalk, draw=white] (2 * 4,0) rectangle (3 * 4, 1) node[pos=0.5] {Sidewalk};
	\draw[white, fill=building, draw=white] (3 * 4,0) rectangle (4 * 4, 1) node[pos=0.5] {Building};
	\draw[white, fill=wall, draw=white] (4 * 4,-0) rectangle (5 * 4, 1) node[pos=0.5] {Wall};
	\draw[black, fill=fence, draw=white] (5 * 4,-0) rectangle (6 * 4, 1) node[pos=0.5] {Fence};
	\draw[white, fill=pole, draw=white] (6 * 4,-0) rectangle (7 * 4, 1) node[pos=0.5] {Pole};
	\draw[white, fill=traffic light, draw=white] (7 * 4,-0) rectangle (8 * 4, 1) node[pos=0.5] {Traffic Light};
	\draw[black, fill=traffic sign, draw=white] (8 * 4,-0) rectangle (9 * 4, 1) node[pos=0.5] {Traffic Sign};
	\draw[white, fill=vegetation, draw=white] (9 * 4,-0) rectangle (10 * 4, 1) node[pos=0.5] {Vegetation};
	\draw[black, fill=terrain, draw=white] (0 * 4,-1) rectangle (1 * 4, 0) node[pos=0.5] {Terrain};
	\draw[white, fill=sky, draw=white] (1 * 4,-1) rectangle (4 * 2, 0) node[pos=0.5] {Sky};
	\draw[white, fill=person, draw=white] (2 * 4,-1) rectangle (3 * 4, 0) node[pos=0.5] {Person};
	\draw[white, fill=rider, draw=white] (3 * 4,-1) rectangle (4 * 4, 0) node[pos=0.5] {Rider};
	\draw[white, fill=car, draw=white] (4 * 4,-1) rectangle (5 * 4, 0) node[pos=0.5] {Car};
	\draw[white, fill=truck, draw=white] (5 * 4,-1) rectangle (6 * 4, 0) node[pos=0.5] {Truck};
	\draw[white, fill=bus, draw=white] (6 * 4,-1) rectangle (7 * 4, 0) node[pos=0.5] {Bus};
	\draw[white, fill=train, draw=white] (7 * 4,-1) rectangle (8 * 4, 0) node[pos=0.5] {Train};
	\draw[white, fill=motorcycle, draw=white] (8 * 4,-1) rectangle (9 * 4, 0) node[pos=0.5] {Motorcycle};
	\draw[white, fill=bicycle, draw=white] (9 * 4,-1) rectangle (10 * 4, 0) node[pos=0.5] {Bicycle};
\end{tikzpicture}}
\caption{The pipeline of our approach for semantic segmentation of nighttime scenes, by transferring knowledge from daytime scenes via the bridge of twilight time scenes.}
\vspace{-2mm}
\label{fig:pipeline}
\end{figure*}
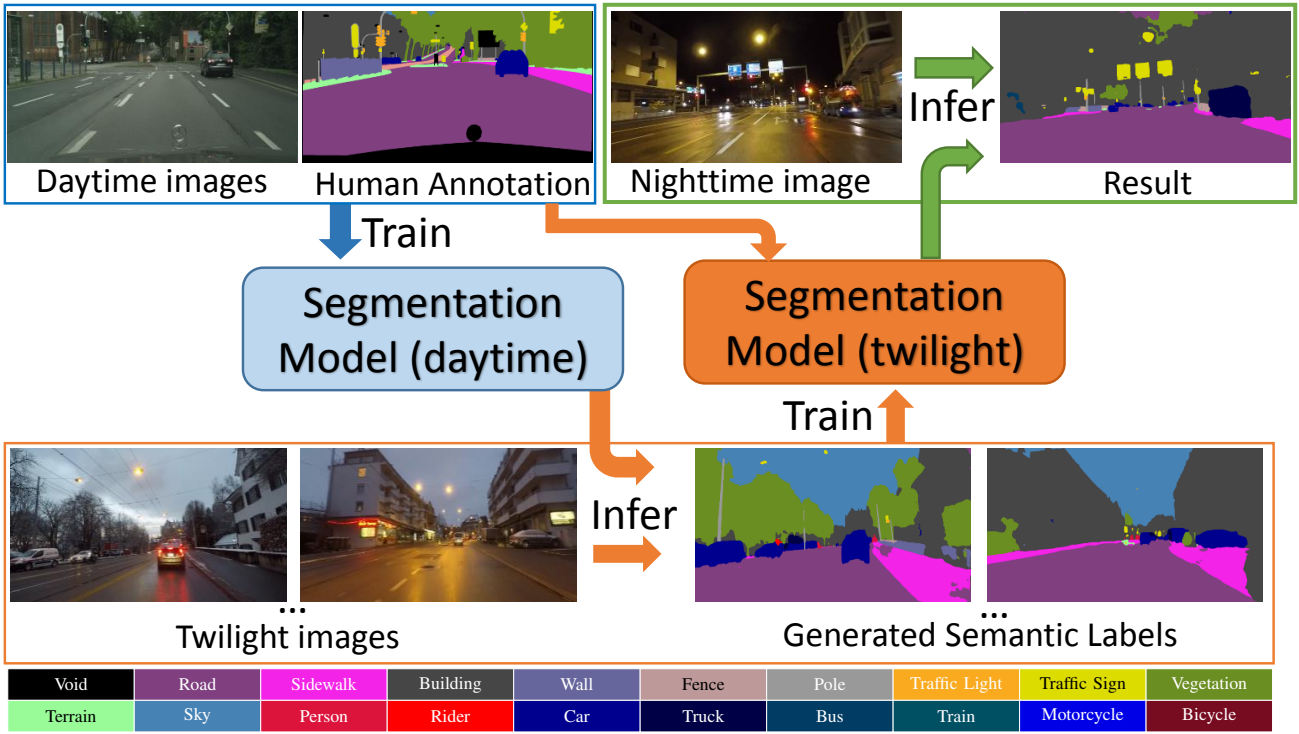

High-level semantic tasks is usually tackled by learning from many annotations of real images. This scheme has achieved a great success for good weather conditions at daytime. Yet, the difficulty of collecting and annotating images for all other weather and illumination conditions renders this standard protocol problematic. To overcome this problem, we depart from this traditional paradigm and propose another route. Instead, we choose to \emph{progressively} adapt the semantic models trained for daytime scenes to nighttime scenes, by using images taken at the twilight time as intermediate stages. The method is based on progressively self-learning scheme, and its  pipeline is shown in Figure \ref{fig:pipeline}. 

The main contributions of the paper are: 1) a novel model adaptation method is developed to transfer semantic knowledge from daytime scenes to nighttime scenes; 2) a new dataset, named \emph{Nighttime Driving}, consisting of images of real  driving scenes at nighttime and twilight time, with $35,000$ unlabeled images and $50$ densely annotated images. These contributions will facilitate the learning and evaluation of semantic segmentation methods for nighttime driving scenes. \emph{Nighttime Driving} is available at \url{http://people.ee.ethz.ch/~daid/NightDriving/}.  

%The paper is organized as follows. Section \ref{sec:related} presents relevant work. Section \ref{sec:approach} is devoted to our method, which is followed by Section \ref{sec:experiment} for the experiments. Finally, Section \ref{sec:conclusion} concludes the paper. 

\section{Related Work} 
\label{sec:related}
%Our work is relevant to nighttime scene understanding, model adaptation, and road scene understanding.
\subsection{Semantic Understanding of Nighttime Scenes}
A lot of work for nighttime object detection/recognition has focused on human detection, by using FIR cameras~\cite{night:vision:pedestrian:05,pedestrian:detection:tracking:night:09} or visible light cameras~\cite{cnn:human:detection:nighttime:17}, or a combination of both~\cite{nighttime:pedestrian:detection:08}. 
There are also notable examples for detecting other road traffic objects such as cars~\cite{nighttime:object:proposal:18} and their rear lights~\cite{night:rear:lights:16}. Another group of work is to develop methods robust to illumination changes for robust road area detection~\cite{road:detection:illumination:invariant} and semantic labeling~\cite{outdoor:transformation:labeling:iv15}. Most of the research in this vein had been conducted before deep learning was widely used. 

Semantic understanding of visual scenes have recently undergone rapid growth, making accurate object detection feasible in images and videos in daytime scenes~\cite{DomainAdaptiveFasterRCNN,refinenet}. It is natural to raise the question of how to extend those sophisticated methods to other weather conditions and illumination conditions, and examine and improve the performance therein. A recent effort has been made for foggy weather~\cite{semantic:foggy:scene}. This work would like to initiate the same research effort for nighttime.   

\subsection{Model Adaptation}
\sloppy{Our work bears resemblance to works from the broad field of transfer learning. Model adaptation across weather conditions to semantically segment simple road scenes is studied in~\cite{road:scene:2013}. 
More recently, domain adaptation based approach was proposed to adapt semantic segmentation models from synthetic images to real environments~\cite{curriculum:domain:adaptation:iccv17,cyCADA,learning:synthetic:data:cvpr18,semantic:foggy:scene,chen2018road}. 
The supervision transfer from daytime scenes to  nighttime scenes in this paper is inspired by the stream of work on model distillation/imitation~\cite{hinton2015distilling,supervision:transfer,dai:metric:imitation}. Our approach is similar in that knowledge is transferred from one domain to the next by distilled from the previous domain.}  The concurrent work in \cite{SynRealDataFogECCV18} on adaptation of semantic
models from clear weather condition to light fog then to dense fog is closely related to ours.
%Combining our method and the aforementioned domain adversarial based methods is a promising direction for future work. 

\subsection{Road Scene Understanding}
Road scene understanding is a crucial enabler for applications such as assisted or autonomous driving. Typical examples include the detection of roads~\cite{recent:progress:lane}, traffic lights~\cite{traffic:light:survey:16}, cars and pedestrians~\cite{Cityscapes,semantic:foggy:scene}, and tracking of such objects~\cite{vehicles:road:survey:13,pathtrack}. We refer the reader to the excellent surveys~\cite{looking:at:human}. The aim of this work is to extend/adapt the advanced models developed recently for road scene understanding at daytime to nighttime, without manually annotating nighttime images. 

\section{Approach} 
\label{sec:approach}
\begin{figure}[t]
\centering
\includegraphics[width=0.9\linewidth]{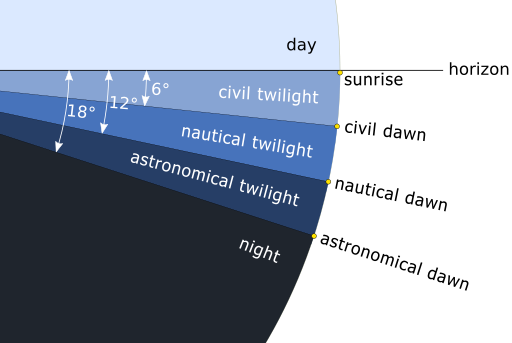}
\caption{Twilight is defined according to the solar elevation angle and is categorized into three subcategories: civil twilight, nautical twilight, and astronomical twilight. (picture is from wikipedia).}
\label{fig:twillight}
\end{figure}

Training a segmentation model with large amount of human annotations should work for nighttime images, similar to what has been achieved for daytime scene understanding \cite{MastRCNN,refinenet}. However, applying this protocol to other weather conditions and illumination conditions is problematic as it is hardly affordable to annotate the same amount of data for all different conditions and their combinations. We depart from this protocol and investigate an automated approach to transfer the knowledge from existing annotations of daytime scenes to nighttime scenes. The approach leverages the fact that illumination changes continuously between daytime and nighttime, through the twilight time. Twilight is the time between dawn and sunrise, or between sunset and dusk. Twilight is defined according to the solar elevation angle, which is the position of the geometric center of the sun relative to the horizon \cite{twilight:definition}.  See Figure \ref{fig:twillight} for an illustration.   

During a large portion of twilight time, solar illumination suffices enough for cameras to capture the terrestrial objects and suffices enough to alleviate the interference of artificial lights to a limited amount. See Figure \ref{fig:pipeline} for examples of road scenes at twilight time. These observations lead to our conjecture that the domain discrepancy between daytime scenes and twilight scenes, and the the domain discrepancy between twilight scenes and nighttime scenes are both smaller than the domain discrepancy between daytime scenes and nighttime scenes. Thus, images captured during twilight time can serve our purpose well --- transfer knowledge from daytime to nighttime. That is, twilight time constructs a bridge for knowledge transfer from our source domain daytime to our target domain nighttime. 

In particular, we train a semantic segmentation model on daytime images using the standard supervised learning paradigm, and apply the model to a large dataset recorded at civil twilight time to generate the class responses. The three subgroups of twilight are used: civil twilight, nautical twilight, and astronomical twilight \cite{twilight:definition}.  Since the domain gap between daytime condition and civil twilight condition is relatively small, these class responses, along with the images, can then be used to fine-tune the semantic segmentation model so that it can adapt to civil twilight time. The same procedure is continued through nautical twilight and astronomical twilight. We then apply the final fine-tuned model to nighttime images. 

This learning approach is inspired by the stream of work on model distillation~\cite{hinton2015distilling,dai:metric:imitation,supervision:transfer}. Those methods either transfer supervision from sophisticated models to simpler models for efficiency~\cite{hinton2015distilling,dai:metric:imitation}, or transfer supervision from the domain of images to other domains such as depth maps~\cite{supervision:transfer}. We here transfer the semantic knowledge of annotations of daytime scenes to nighttime scenes via  the unlabeled images recorded at twilight time. 

Let us denote an image by $\mathbf{x}$, and indicate the image taken at \emph{daytime}, \emph{civil twilight time}, \emph{nautical twilight time}, \emph{astronomical twilight time} and \emph{nighttime} by  $\mathbf{x}^0$, $\mathbf{x}^1$, $\mathbf{x}^2$, $\mathbf{x}^3$, and $\mathbf{x}^4$, respectively. 
The corresponding human annotation for $\mathbf{x}^0$ is provided and denoted by $\mathbf{y}^0$,  where $\mathbf{y}^0(m,n) \in\{1, ..., C\}$  is the label of pixel $(m,n)$, and $C$ is the total number of classes. Then, the training data consist of labeled data at daytime $\mathcal{D}^0 =\{(\mathbf{x}^0_i, \mathbf{y}^0_{i})\}_{i=1}^{l^0}$, and three unlabeled datasets for the three twilight categories: $\mathcal{D}^1=\{\mathbf{x}^1_{j}\}_{j=1}^{l^1}$, $\mathcal{D}^2=\{\mathbf{x}^2_{k}\}_{k=1}^{l^2}$, and $\mathcal{D}^3=\{\mathbf{x}^3_{q}\}_{q=1}^{l^3}$, where $l^0$, $l^1$, $l^2$, and $l^3$ are the total number of images in the corresponding datasets.  
% \begin{algorithm}
% \caption{Progressive Model Adaptation}\label{alg:1}
% \begin{algorithmic}[1]
% \State $\textit{stringlen} \gets 
% \min_{\phi^{\prime \prime}} \frac{1}{l}\sum_{i=1}^l L(\phi^{\prime \prime}(\mathbf{x}^{\prime \prime}_i), \mathbf{y}_i) + \lambda \frac{1}{u}\sum_{j=l+1}^{l+u} L(\phi^{\prime \prime}(\mathbf{\hat{x}}^\prime_j), \hat{\mathbf{y}}^\prime_j),
% \label{eq:ssl}
%  $
% \State $i \gets \textit{patlen}$
% \BState \emph{top}:
% \If {$i > \textit{stringlen}$} \Return false
% \EndIf
% \State $j \gets \textit{patlen}$
% \BState \emph{loop}:
% \If {$\textit{string}(i) = \textit{path}(j)$}
% \State $j \gets j-1$.
% \State $i \gets i-1$.
% \State \textbf{goto} \emph{loop}.
% \State \textbf{close};
% \EndIf
% \State $i \gets i+\max(\textit{delta}_1(\textit{string}(i)),\textit{delta}_2(j))$.
% \State \textbf{goto} \emph{top}.
% \end{algorithmic}
% \end{algorithm}

The method consists of eight steps and it is summarized below. 
\begin{enumerate}
\item [\textbf{1}:]  train a segmentation model with daytime images and the human annotations: 
\begin{equation}
\min_{\phi^0} \frac{1}{l^0}\sum_{i=1}^{l^0} L(\phi^0(\mathbf{x}^0_i), \mathbf{y}^0_i),
\end{equation} where $L(.,.)$ is the cross entropy loss function;  \label{item1} 
\item [\textbf{2}:] apply  segmentation model $\phi^0$ to the images recorded at civil twilight time to obtain ``noisy'' semantic labels:  $\hat{\mathbf{y}}^1 = \phi^0(\mathbf{x}^1)$, and augment dataset $\mathcal{D}^1$ to $\hat{\mathcal{D}}^1$: $\hat{\mathcal{D}}^1=\{(\mathbf{x}^1_j, \hat{\mathbf{y}}^1_j)\}_{j=1}^{l^1}$;  \label{item2} 
\item [\textbf{3}:] instantiate a new model $\phi^1$ by duplicating $\phi^0$, and then fine-tune (retrain) the semantic model on $\mathcal{D}^0$ and $\hat{\mathcal{D}}^1$: 
\begin{equation}
\phi^1 \leftarrow \phi^0,
\end{equation}
and  
\begin{equation}
\min_{\phi^1} \Big(\frac{1}{l^0}\sum_{i=1}^{l^0} L(\phi^1(\mathbf{x}^0_i), \mathbf{y}^0_i) + \frac{\lambda^1 }{l^1}\sum_{j=1}^{l^1} L(\phi^1(\mathbf{x}^1_j), \hat{\mathbf{y}}^1_j) \Big),
\label{eq:step3}
\end{equation}
where $\lambda^1$ is a hyper-parameter balancing the weights of the two data sources; \label{item3}
\item [\textbf{4}:] apply segmentation model $\phi^1$ to the images recorded at nautical twilight time to obtain ``noisy'' semantic labels:  $\hat{\mathbf{y}}^2 = \phi^1(\mathbf{x}^2)$, and augment dataset $\mathcal{D}^2$ to $\hat{\mathcal{D}}^2$: $\hat{\mathcal{D}}^2=\{(\mathbf{x}^2_k, \hat{\mathbf{y}}^2_k)\}_{k=1}^{l^2}$; 
\label{item4} 
\item [\textbf{5}:] instantiate a new model $\phi^2$ by duplicating $\phi^1$, and fine-tune (train) semantic model on $\mathcal{D}^0$, $\hat{\mathcal{D}}^1$ and $\hat{\mathcal{D}}^2$: 
\begin{equation}
\phi^2 \leftarrow \phi^1,
\end{equation}
and then 
\begin{multline}
\min_{\phi^2} \Big( \frac{1}{l^0}\sum_{i=1}^{l^0} L(\phi^2(\mathbf{x}^0_i), \mathbf{y}^0_i) +  \frac{\lambda^1}{l^1}\sum_{j=1}^{l^1} L(\phi^2(\mathbf{x}^1_j), \hat{\mathbf{y}}^1_j) \\ 
+  \frac{\lambda^2}{l^2}\sum_{k=1}^{l^2} L(\phi^2(\mathbf{x}^2_k), \hat{\mathbf{y}}^2_k)   \Big),
\label{eq:step5}
\end{multline}
where $\lambda^1$ and $\lambda^2$ are  hyper-parameters regulating the weights of the datasets; \label{item5}
\item [\textbf{6}:] apply segmentation model $\phi^2$ to the images recorded at astronomical twilight data to obtain ``noisy'' semantic labels:  $\hat{\mathbf{y}}^3 = \phi^2(\mathbf{x}^3)$, and augment dataset $\mathcal{D}^3$ to $\hat{\mathcal{D}}^3$: $\hat{\mathcal{D}}^3=\{(\mathbf{x}^3_q, \hat{\mathbf{y}}^3_q)\}_{q=1}^{l^3}$; 
;  \label{item6} 
\item [\textbf{7}:] instantiate a new model $\phi^3$ by duplicating $\phi^2$, and fine-tune (train) the semantic model on all four datasets $\mathcal{D}^0$, $\hat{\mathcal{D}}^1$, $\hat{\mathcal{D}}^2$ and $\hat{\mathcal{D}}^3$: 
\begin{equation}
\phi^3 \leftarrow \phi^2,
\end{equation}
and then 
\begin{multline}
\min_{\phi^3} \Big( \frac{1}{l^0}\sum_{i=1}^{l^0} L(\phi^3(\mathbf{x}^0_i), \mathbf{y}^0_i) +  \frac{\lambda^1}{l^1}\sum_{j=1}^{l^1} L(\phi^3(\mathbf{x}^1_j), \hat{\mathbf{y}}^1_j) \\ 
+  \frac{\lambda^2}{l^2}\sum_{k=1}^{l^2} L(\phi^3(\mathbf{x}^2_k), \hat{\mathbf{y}}^2_k) + \frac{\lambda^3}{l^3}\sum_{q=1}^{l^3} L(\phi^3(\mathbf{x}^3_q), \hat{\mathbf{y}}^3_q)   \Big),
\label{eq:step7}
\end{multline}
where $\lambda^1$, $\lambda^1$ and $\lambda^3$ are hyper-parameters regulating the weights of the datasets; 
\label{item7}
\item [\textbf{8}:] apply model $\phi^3$ to nighttime images to perform the segmentation: $\hat{\mathbf{y}}^4 = \phi^3(\mathbf{x}^4)$.
\end{enumerate}

We term our method Gradual Model Adaptation. During training, in order to balance the weights of different data sources (in Equation~\ref{eq:step3}, Equation~\ref{eq:step5} and Equation~\ref{eq:step7}), we empirically give equal weight to all training datasets. An optimal value can be obtained via cross-validation. The optimization of Equation~\ref{eq:step3}, Equation~\ref{eq:step5} and Equation~\ref{eq:step7} are implemented by feeding to the training algorithm a stream of hybrid data, for which images in the considered datasets are sampled proportionally according to the parameters $\lambda^1$, $\lambda^2$, and $\lambda^3$. In this work, they all set to $1$, which means all datasets are sampled at the same rate.

Rather than applying the model trained on daytime images directly to nighttime images, Gradual Model Adaptation breaks down the problem to three progressive steps to adapt the semantic model. In each of the step, the domain gap is much smaller than the domain gap between daytime domain and nighttime domain. Due to the unsupervised nature of this domain adaptation, the algorithm will also be affected by the noise in the labels. The daytime dataset $\mathcal{D}^1$ is always used for the training, to balance between noisy data of similar domains and clean data of a distinct domain.

\section{Experiments} 
\label{sec:experiment} 

\subsection{Data Collection}
\emph{Nighttime Driving} was collected during 5 rides with a car inside multiple Swiss cities and their suburbs using a GoPro~Hero~5 camera. We recorded 5 large video sequence with length of about 2 hours. The video recording starts from daytime, goes through twilight time and ends at full nighttime. The video frames are extracted at a rate of one frame per second, leading to 35,000 images in total. According to \cite{twilight:definition} and the sunset time of each recording day, we partition the dataset into five parts: daytime, civil twilight time, nautical twilight time, astronomical twilight time, and nighttime. They consist of 8000, 8750, 8750, 8750, and 9500 images, respectively.  

We manually select 50 nighttime images of diverse visual scenes, and construct the test set of \emph{Nighttime Driving} therefrom, which we term \emph{Nighttime Driving-test}. 
The aforementioned selection is performed manually in order to guarantee that the test set has high diversity, which compensates for its relatively small size in terms of statistical significance of evaluation results. 
We annotate these images with fine pixel-level semantic annotations using the 19 evaluation classes of the Cityscapes dataset~\cite{Cityscapes}: \emph{road}, \emph{sidewalk}, \emph{building}, \emph{wall}, \emph{fence}, \emph{pole}, \emph{traffic light}, \emph{traffic sign}, \emph{vegetation}, \emph{terrain}, \emph{sky}, \emph{person}, \emph{rider}, \emph{car}, \emph{truck}, \emph{bus}, \emph{train}, \emph{motorcycle} and \emph{bicycle}. 
In addition, we assign the \emph{void} label to pixels which do not belong to any of the above 19 classes, or the class of which is uncertain due to insufficient illumination. Every such pixel is ignored for semantic segmentation evaluation. 
%Comprehensive statistics for the semantic annotations of \emph{Nighttime Driving-test} are presented in Figure~\ref{fig:dataset:stats}. 

\subsection{Experimental Evaluation}
Our model of choice for experiments on semantic segmentation is the RefineNet~\cite{refinenet}. We use the publicly available \emph{RefineNet-res101-Cityscapes} model, which has been trained on the daytime training set of Cityscapes. In all experiments of this section, we use a constant base learning rate of $5\times{}10^{-5}$ and mini-batches of size 1.  

Our segmentation experiment showcases the effectiveness of our model adaptation pipeline, using twilight time as a bridge. The models which are obtained after the initial adaptation step are further fine-tuned on the union of the daytime Cityscapes dataset and the previously segmented twilight datasets, where the latter sets are labeled by the adapted models one step ahead. 

We evaluate four variants of our method and compare them to the original segmentation model trained on daytime images directly. Using the pipeline described in Section \ref{sec:approach}, three models can be obtained, in particular $\phi^1$, $\phi^2$, and $\phi^3$. 

We also compare to an alternative adaptation approach which generates labels (by using the original model trained on daytime data) for all twilight images at once and fine-tunes the original daytime segmentation model once. To put in another word, the three-step progressive model adaptation is reduced to a one-step progressive model adaptation.  

\begin{table}[!tb]
  \centering
  \caption{Performance comparison between the variants of our method to the original segmentation model.}
  \label{table:experiments}
  \setlength\tabcolsep{5pt}
    \begin{tabular}{lcc}
Model   & Fine-tuning on twilight data  & Mean IoU \\
\toprule
Refinenet \cite{refinenet} & --- & 35.2 \\
\midrule
Refinenet  & $\phi^1$ ($\rightarrow$ civil) & 38.6 \\
Refinenet  & $\phi^2$ ($\rightarrow$  civil  $\rightarrow$  nautical) & 39.9 \\
Refinenet  & $\phi^3$ ($\rightarrow$ civil  $\rightarrow$  nautical $\rightarrow$ astronomical) & \textbf{41.6} \\
\midrule
Refinenet & $\rightarrow$ all twilight (1-step adaptation) & 39.1 \\
\end{tabular}
\vspace{-2mm}
\end{table}

\textbf{Quantitative Results}. 
The overall intersection over union (IoU) over all classes of the semantic segmentation by all methods are reported in Tables~\ref{table:experiments}. The table shows that all variants of our adaptation method improve the performance of the original semantic model trained with daytime data. This is mainly due to the fact that twilight time fall into the middle ground of daytime and nighttime, so the domain gaps from twilight to the other two domains are smaller than the direct domain gap of the two.

\begin{figure*}[!tb]
\centering
\begin{tabular}{ccccc}
\hspace{-3mm}
\includegraphics[width=0.24\textwidth]{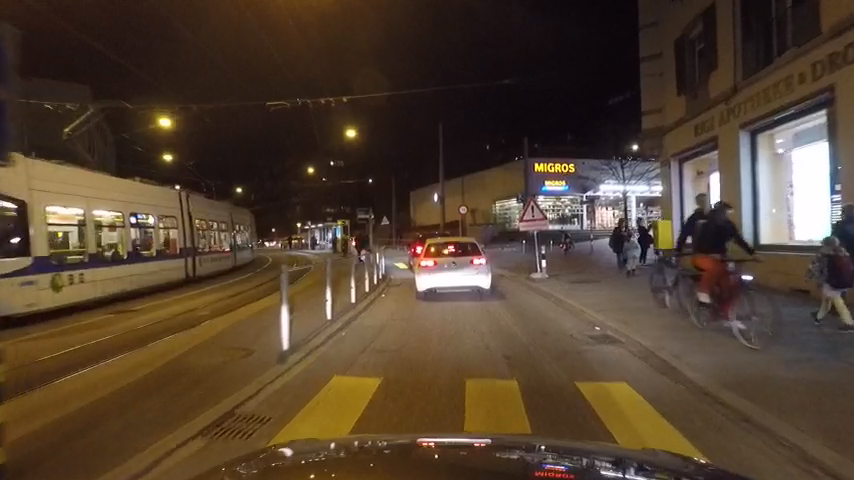} 
& \hspace{-4.5mm}
\includegraphics[width=0.24\textwidth]{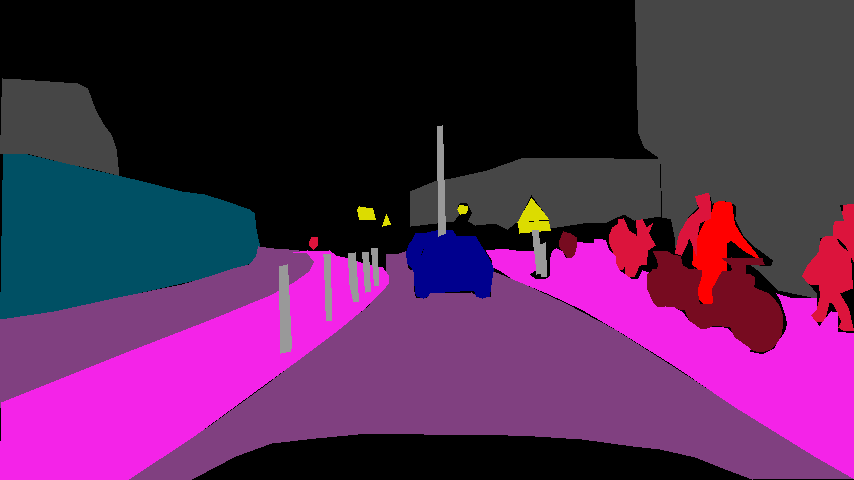}  
& \hspace{-4.5mm}   
\includegraphics[width=0.24\textwidth]{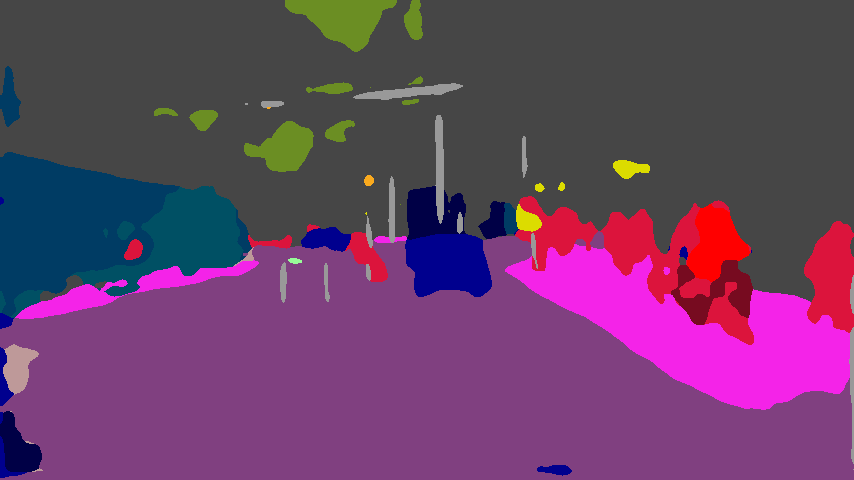} 
& \hspace{-4.5mm} 
\includegraphics[width=0.24\textwidth]{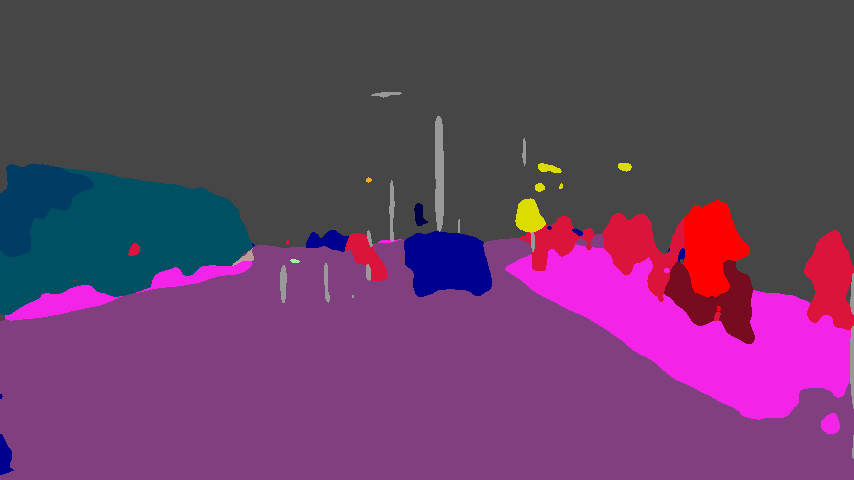} \\ 
\hspace{-3mm}
\includegraphics[width=0.24\textwidth]{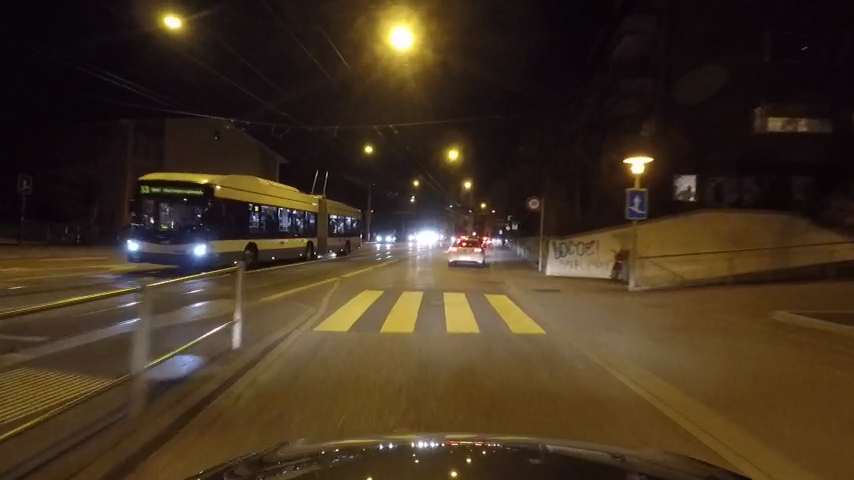} 
& \hspace{-4.5mm}
\includegraphics[width=0.24\textwidth]{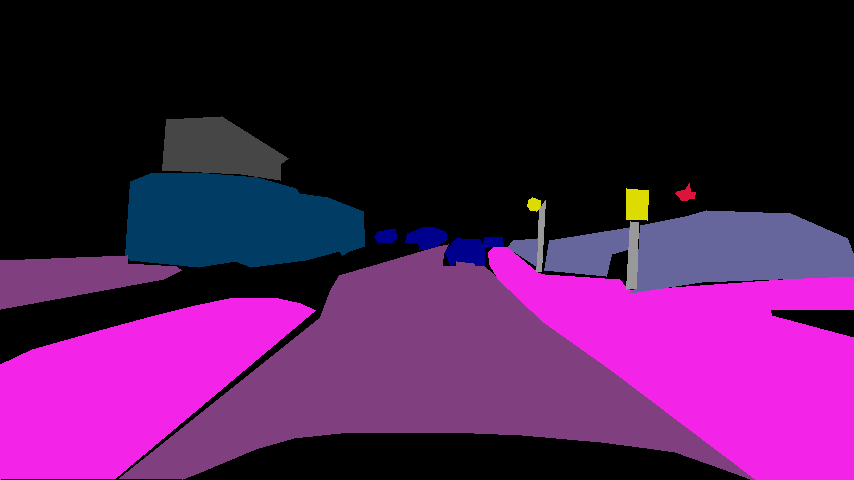}  
& \hspace{-4.5mm}   
\includegraphics[width=0.24\textwidth]{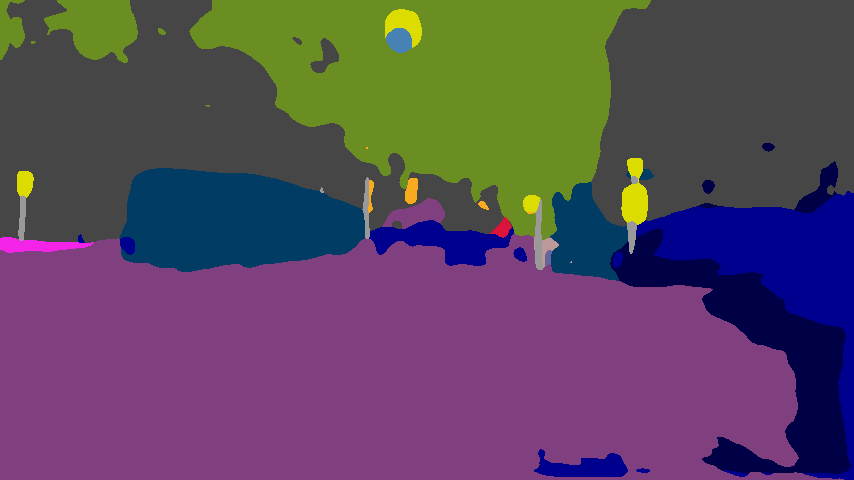} 
& \hspace{-4.5mm} 
\includegraphics[width=0.24\textwidth]{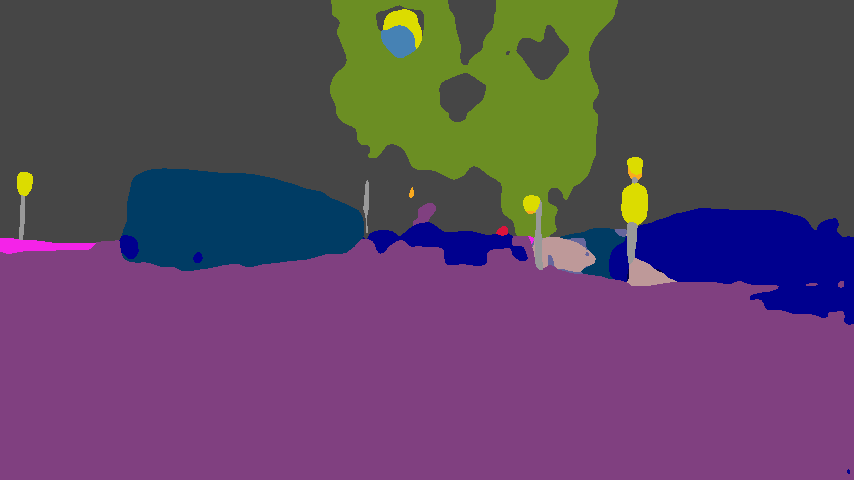} \\ 
\hspace{-3mm}
\includegraphics[width=0.24\textwidth]{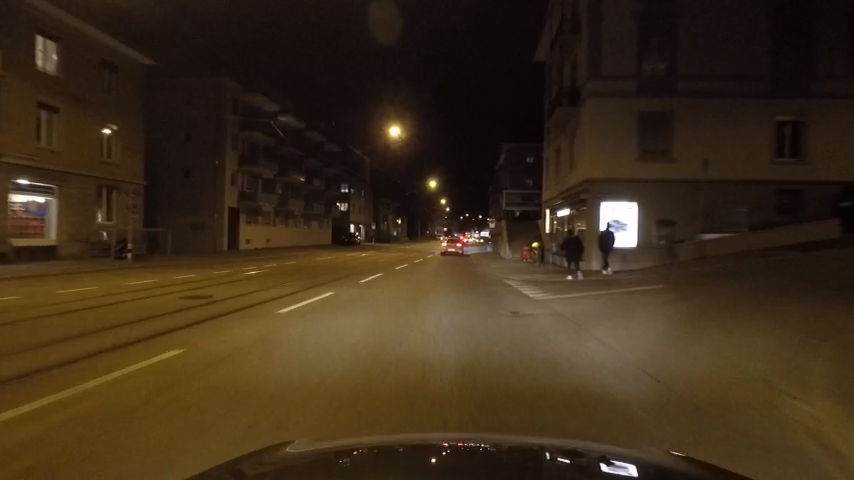} 
& \hspace{-4.5mm}
\includegraphics[width=0.24\textwidth]{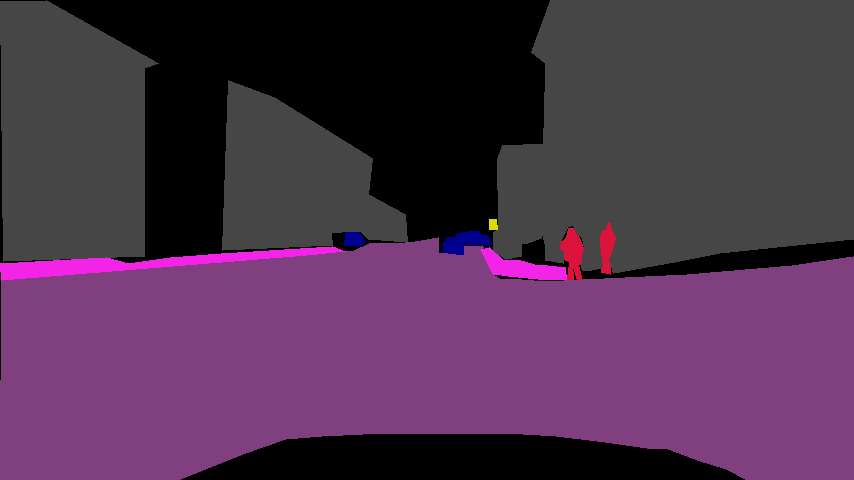}  
& \hspace{-4.5mm}   
\includegraphics[width=0.24\textwidth]{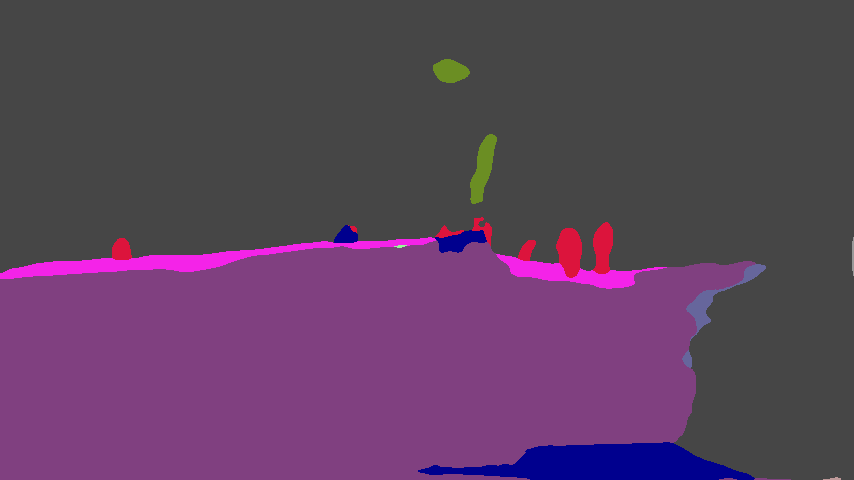} 
& \hspace{-4.5mm} 
\includegraphics[width=0.24\textwidth]{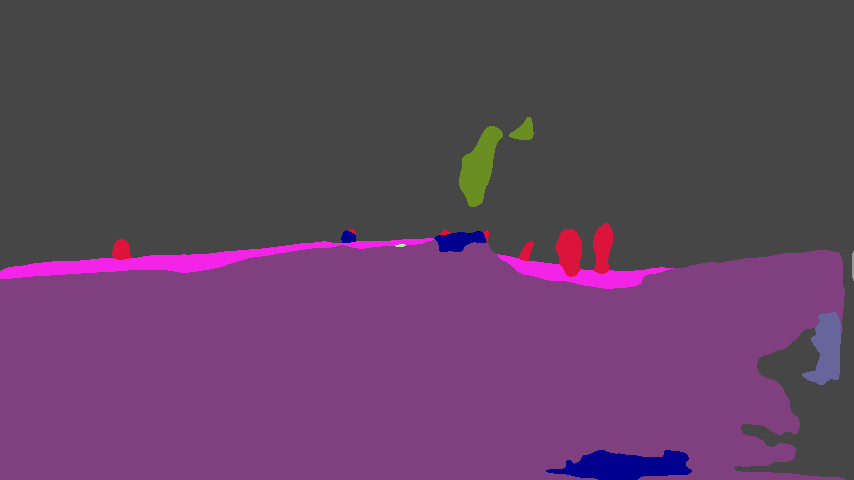} \\ 
\hspace{-3mm}
\includegraphics[width=0.24\textwidth]{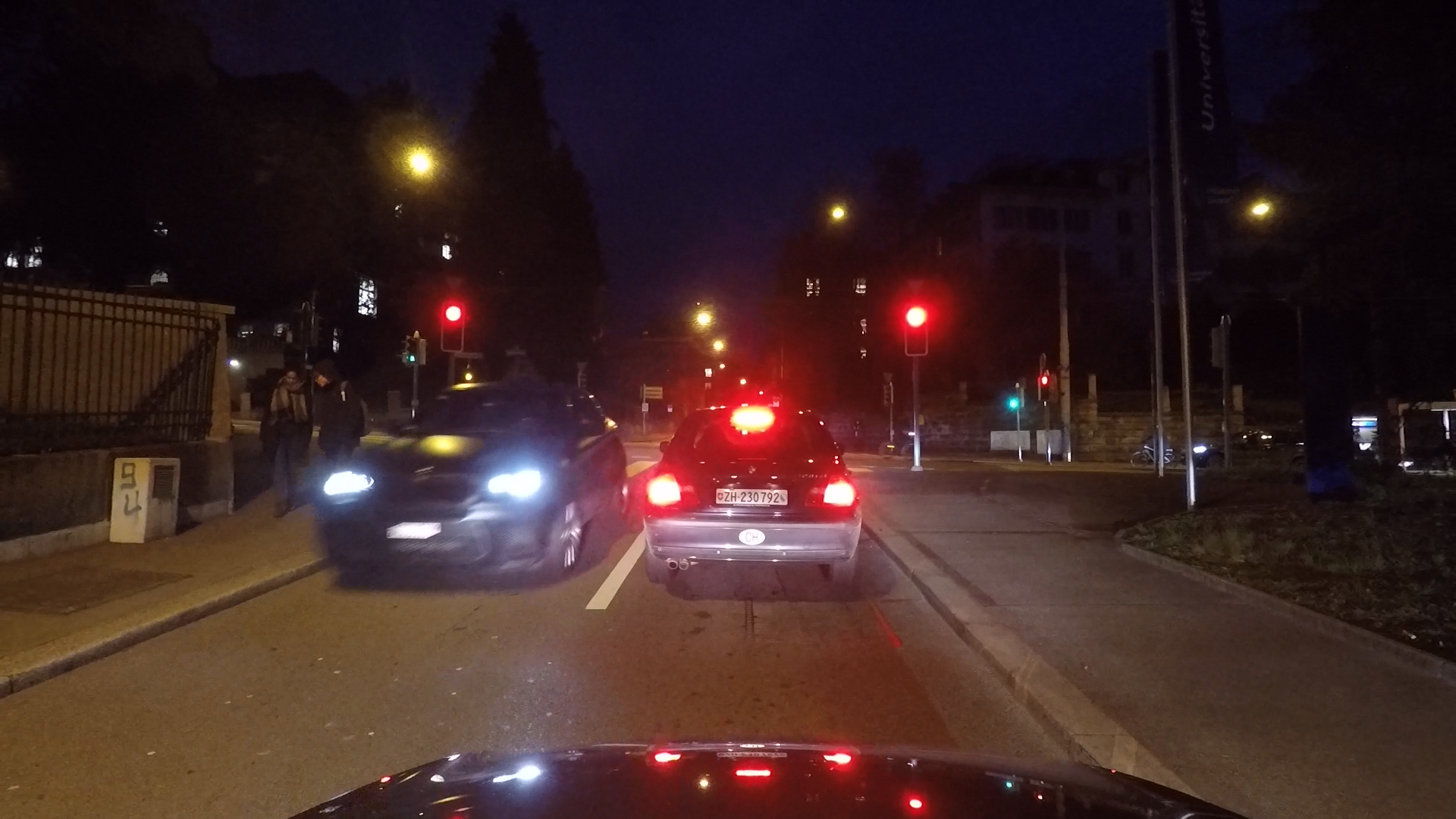} 
& \hspace{-4.5mm}
\includegraphics[width=0.24\textwidth]{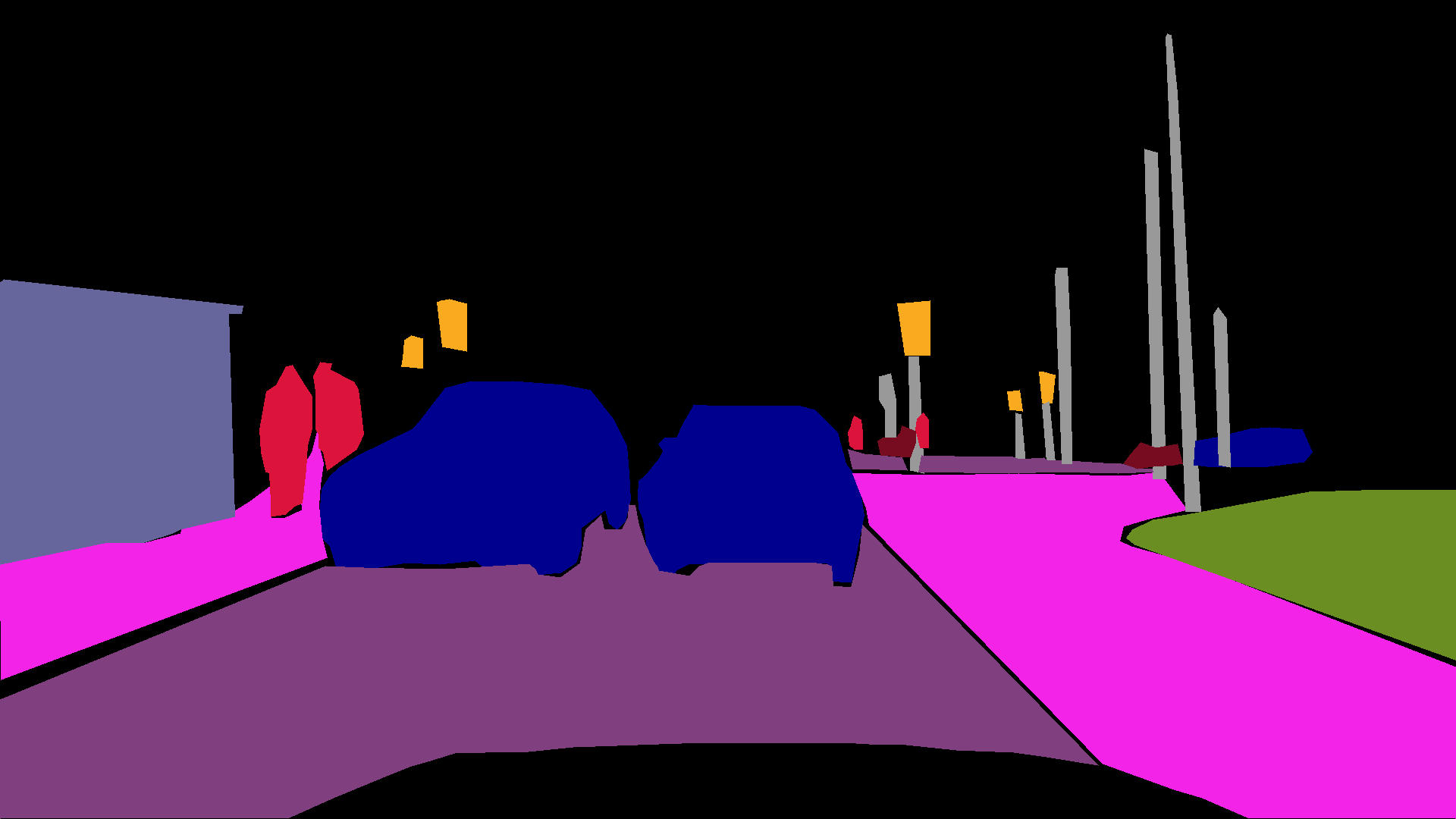}  
& \hspace{-4.5mm}   
\includegraphics[width=0.24\textwidth]{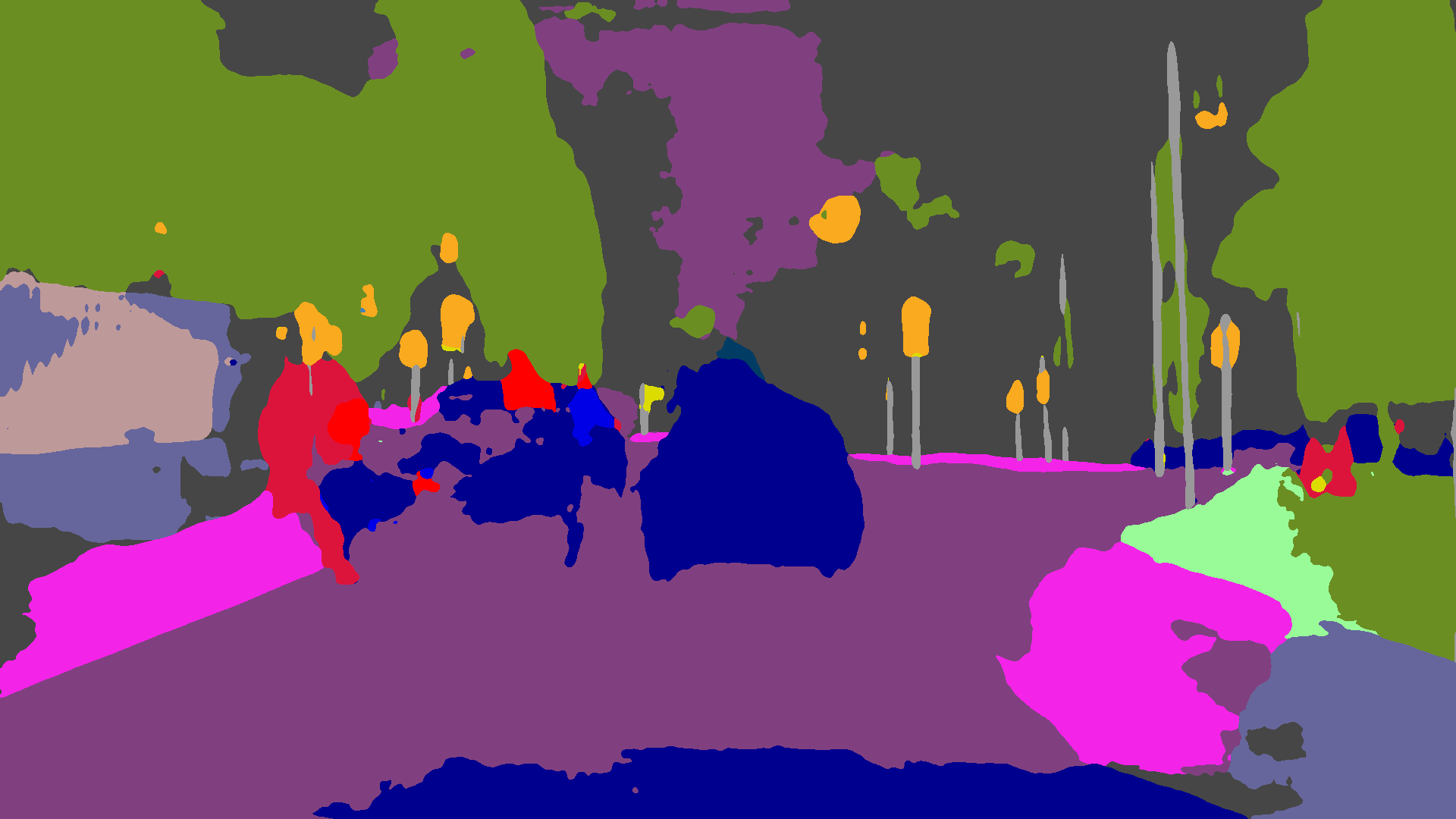} 
& \hspace{-4.5mm} 
\includegraphics[width=0.24\textwidth]{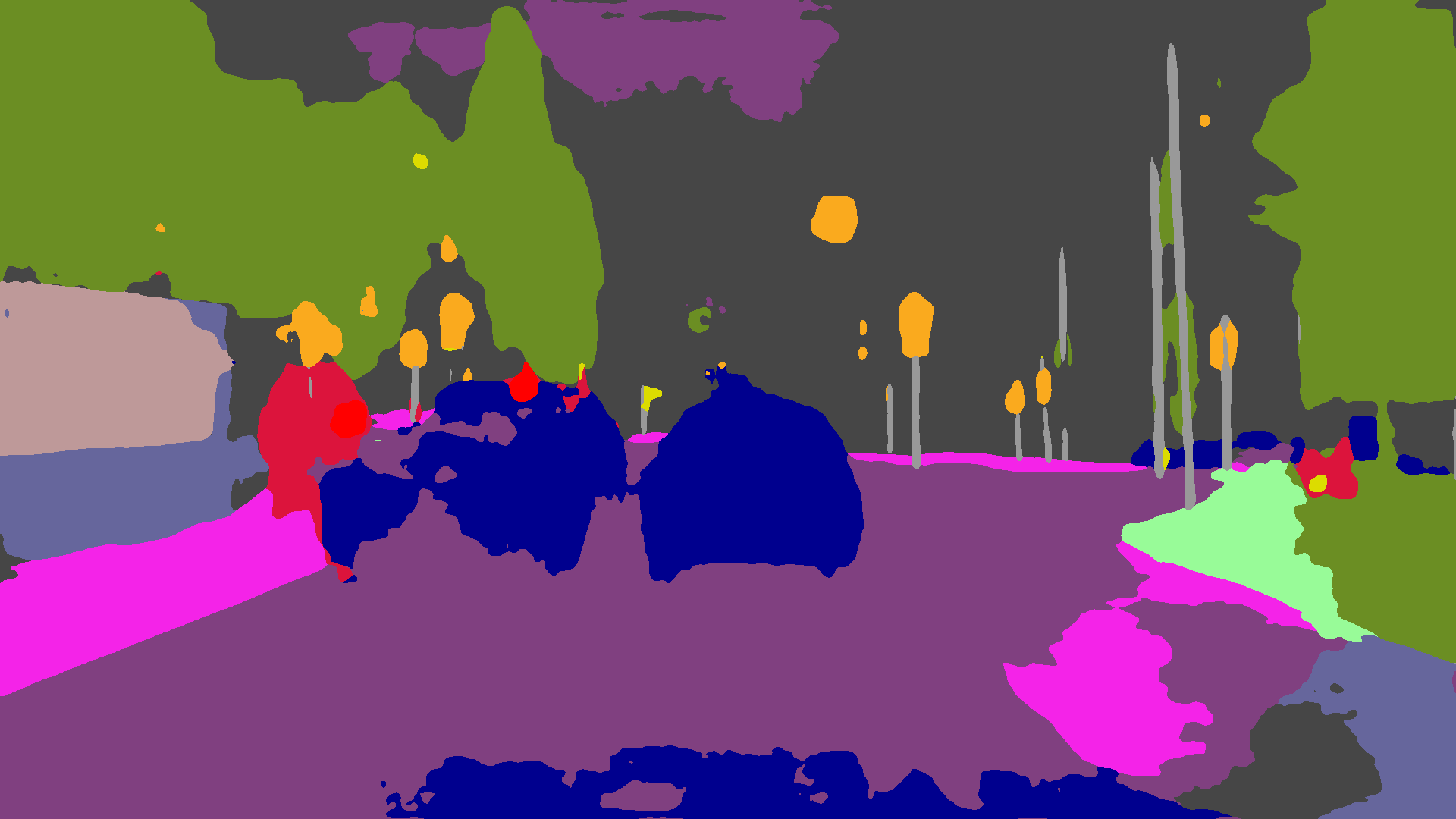} \\ 
\hspace{-3mm}
\includegraphics[width=0.24\textwidth]{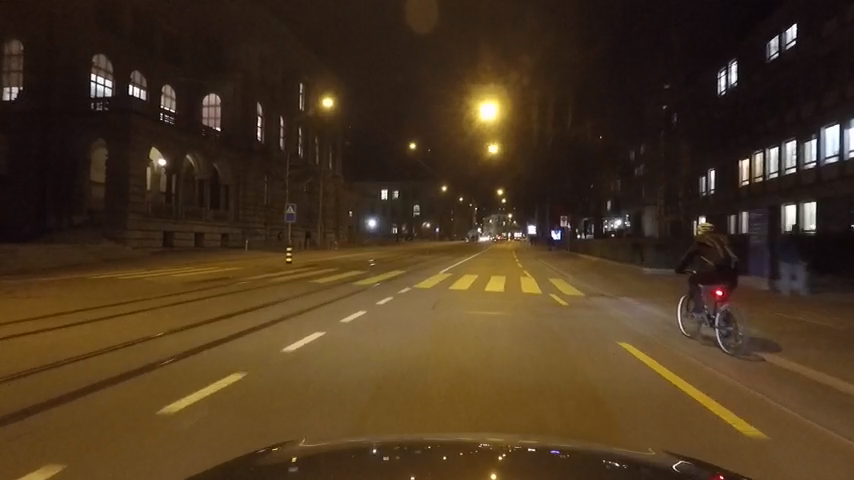} 
& \hspace{-4.5mm}
\includegraphics[width=0.24\textwidth]{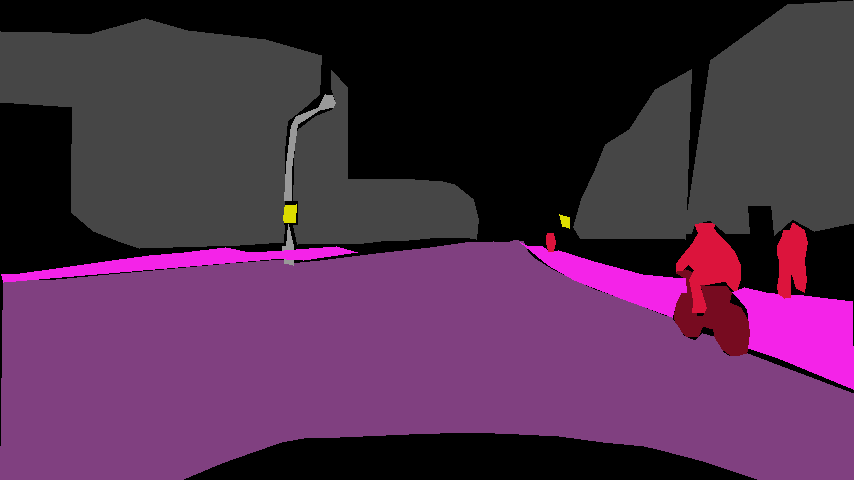}  
& \hspace{-4.5mm}   
\includegraphics[width=0.24\textwidth]{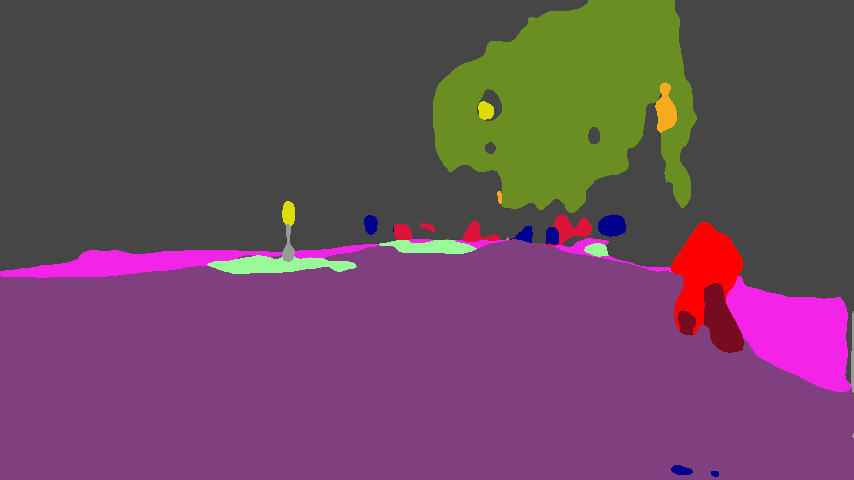} 
& \hspace{-4.5mm} 
\includegraphics[width=0.24\textwidth]{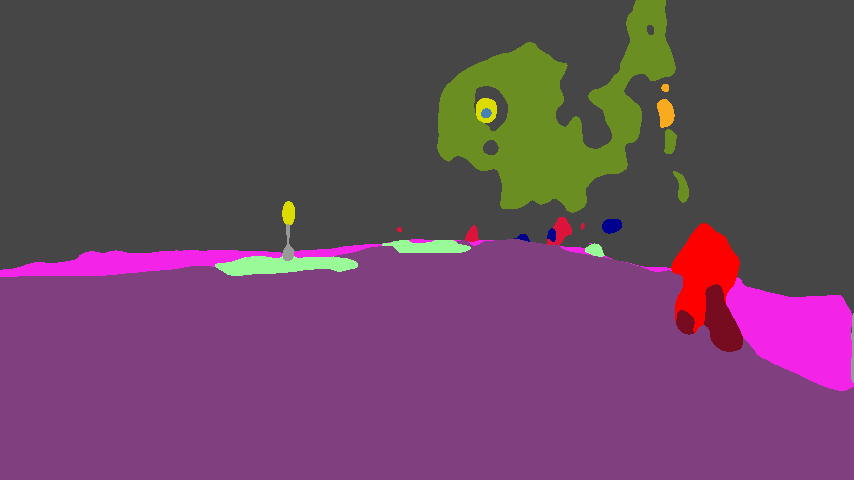} \\ 
\hspace{-3mm}
\includegraphics[width=0.24\textwidth]{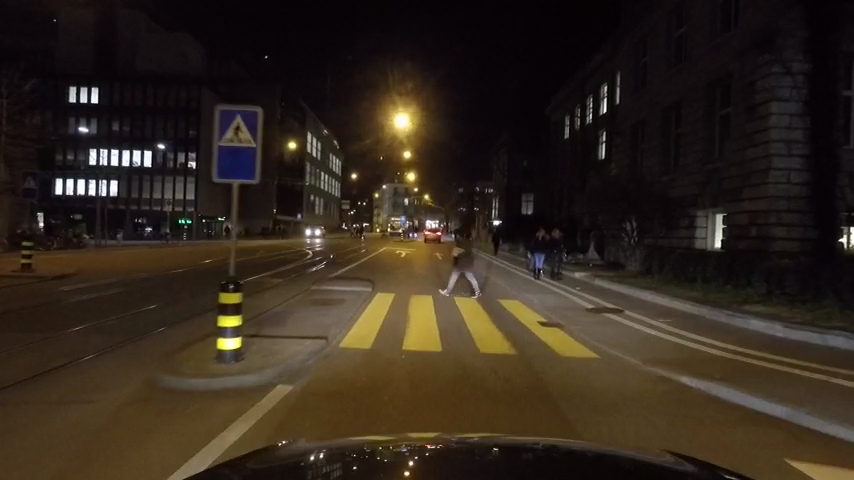} 
& \hspace{-4.5mm}
\includegraphics[width=0.24\textwidth]{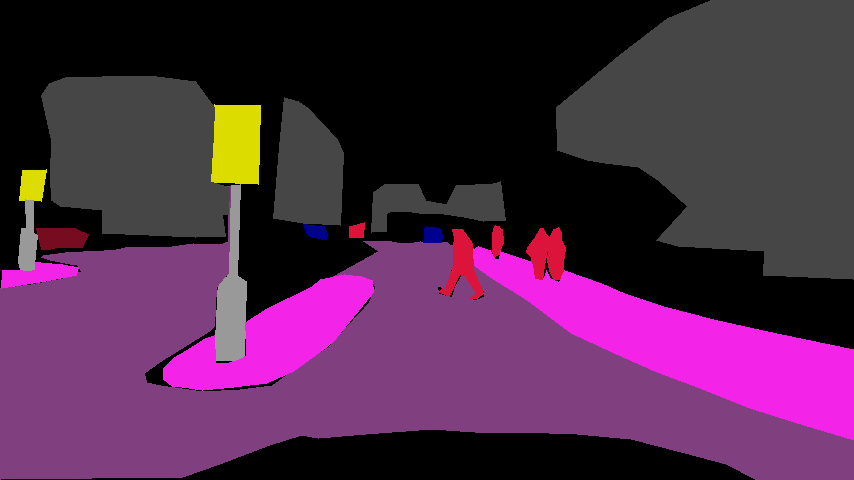}  
& \hspace{-4.5mm}   
\includegraphics[width=0.24\textwidth]{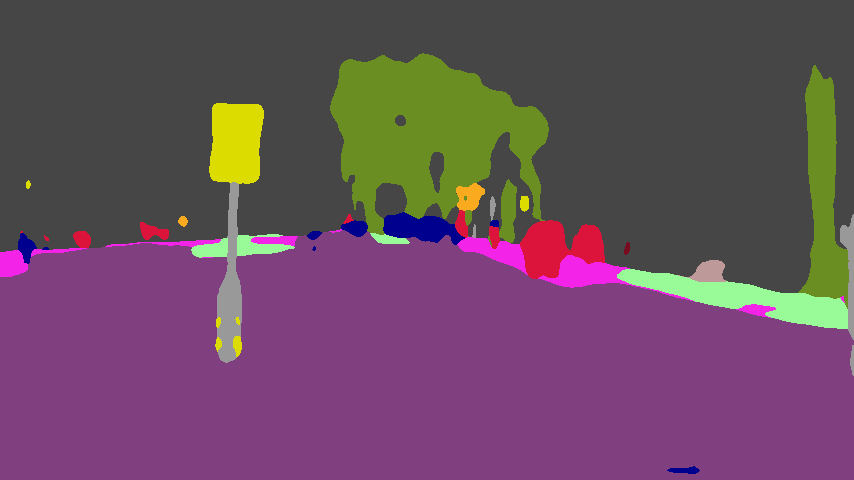} 
& \hspace{-4.5mm} 
\includegraphics[width=0.24\textwidth]{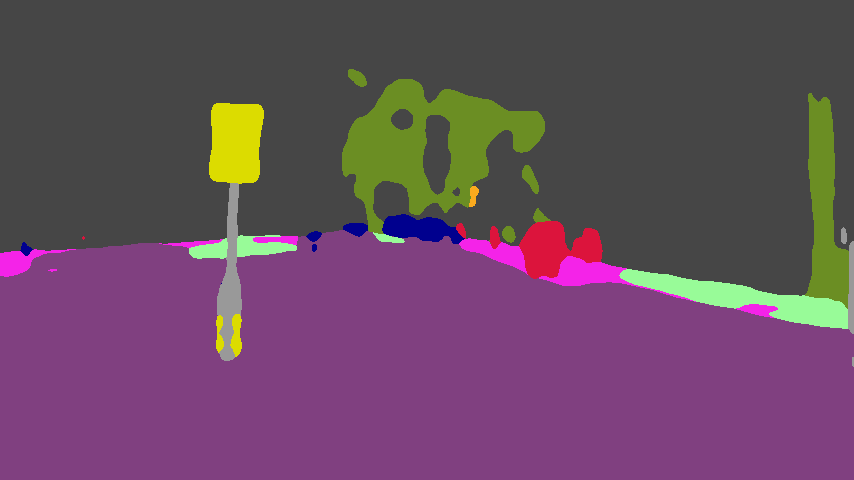} \\ 
\hspace{-3mm}
\includegraphics[width=0.24\textwidth]{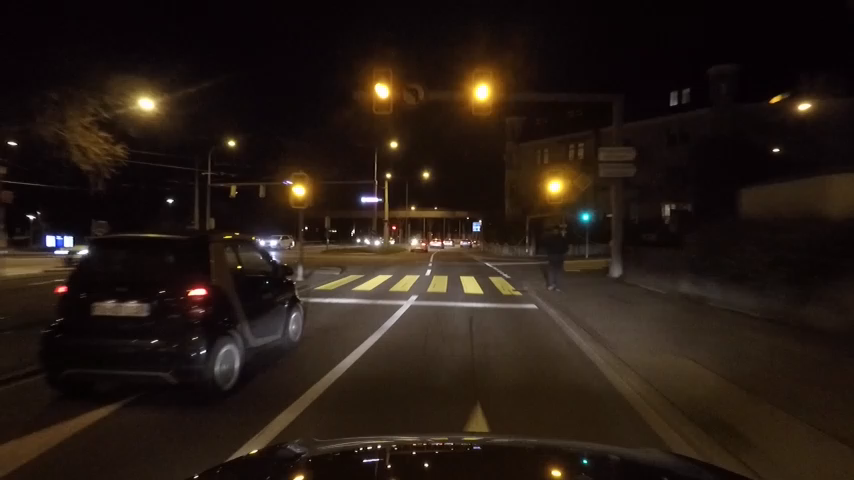} 
& \hspace{-4.5mm}
\includegraphics[width=0.24\textwidth]{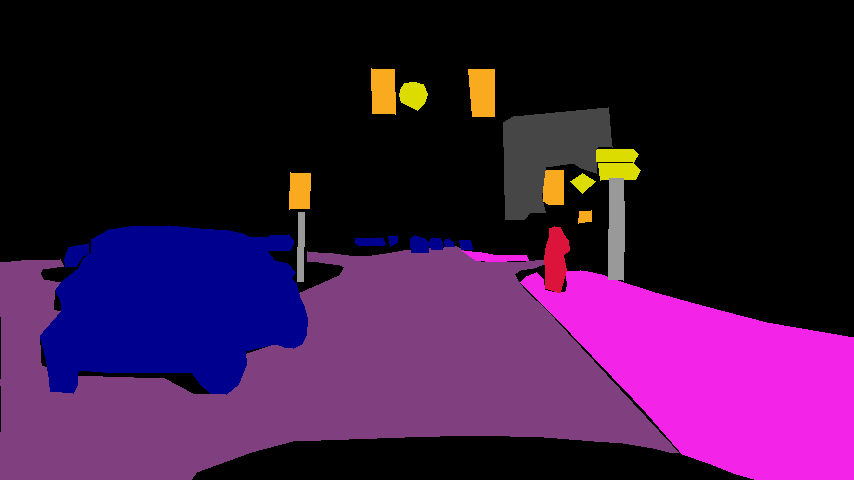}  
& \hspace{-4.5mm}   
\includegraphics[width=0.24\textwidth]{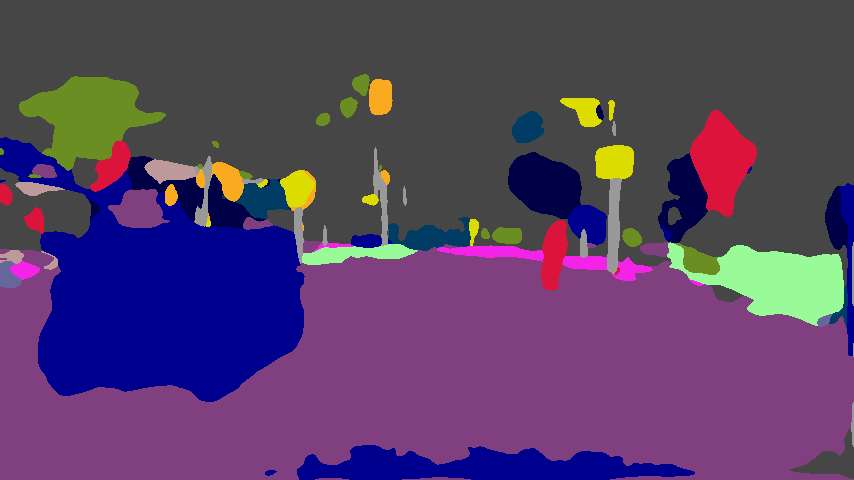} 
& \hspace{-4.5mm} 
\includegraphics[width=0.24\textwidth]{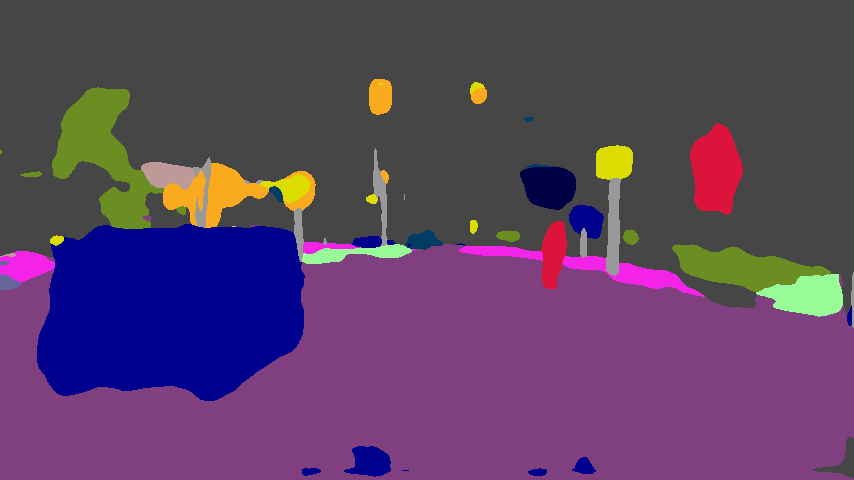} \\
\hspace{-1mm} \text{(a) nighttime image} &  \hspace{-2mm} (b) ground truth &  \hspace{-2mm} \text{(c) refineNet~\cite{refinenet} } &  \hspace{-2mm} \text{(d) our method}  \\
\end{tabular}
\resizebox{\linewidth}{!}{
\begin{tikzpicture}[tight background, scale=0.75, every node/.style={font=\large}]
	\draw[white, fill=void, draw=white] (0,0) rectangle (1 * 4, 1) node[pos=0.5] {Void};
	\draw[white, fill=road, draw=white] (1 * 4,0) rectangle (2 * 4, 1) node[pos=0.5] {Road};
	\draw[white, fill=sidewalk, draw=white] (2 * 4,0) rectangle (3 * 4, 1) node[pos=0.5] {Sidewalk};
	\draw[white, fill=building, draw=white] (3 * 4,0) rectangle (4 * 4, 1) node[pos=0.5] {Building};
	\draw[white, fill=wall, draw=white] (4 * 4,-0) rectangle (5 * 4, 1) node[pos=0.5] {Wall};
	\draw[black, fill=fence, draw=white] (5 * 4,-0) rectangle (6 * 4, 1) node[pos=0.5] {Fence};
	\draw[white, fill=pole, draw=white] (6 * 4,-0) rectangle (7 * 4, 1) node[pos=0.5] {Pole};
	\draw[white, fill=traffic light, draw=white] (7 * 4,-0) rectangle (8 * 4, 1) node[pos=0.5] {Traffic Light};
	\draw[black, fill=traffic sign, draw=white] (8 * 4,-0) rectangle (9 * 4, 1) node[pos=0.5] {Traffic Sign};
	\draw[white, fill=vegetation, draw=white] (9 * 4,-0) rectangle (10 * 4, 1) node[pos=0.5] {Vegetation};
	\draw[black, fill=terrain, draw=white] (0 * 4,-1) rectangle (1 * 4, 0) node[pos=0.5] {Terrain};
	\draw[white, fill=sky, draw=white] (1 * 4,-1) rectangle (4 * 2, 0) node[pos=0.5] {Sky};
	\draw[white, fill=person, draw=white] (2 * 4,-1) rectangle (3 * 4, 0) node[pos=0.5] {Person};
	\draw[white, fill=rider, draw=white] (3 * 4,-1) rectangle (4 * 4, 0) node[pos=0.5] {Rider};
	\draw[white, fill=car, draw=white] (4 * 4,-1) rectangle (5 * 4, 0) node[pos=0.5] {Car};
	\draw[white, fill=truck, draw=white] (5 * 4,-1) rectangle (6 * 4, 0) node[pos=0.5] {Truck};
	\draw[white, fill=bus, draw=white] (6 * 4,-1) rectangle (7 * 4, 0) node[pos=0.5] {Bus};
	\draw[white, fill=train, draw=white] (7 * 4,-1) rectangle (8 * 4, 0) node[pos=0.5] {Train};
	\draw[white, fill=motorcycle, draw=white] (8 * 4,-1) rectangle (9 * 4, 0) node[pos=0.5] {Motorcycle};
	\draw[white, fill=bicycle, draw=white] (9 * 4,-1) rectangle (10 * 4, 0) node[pos=0.5] {Bicycle};
\end{tikzpicture}}
\vspace{-4mm}
\caption{Qualitative results for semantic segmentation on \emph{Nighttime Driving-test}. }
\vspace{-4mm}
\label{fig:sem:seg1}
\end{figure*}
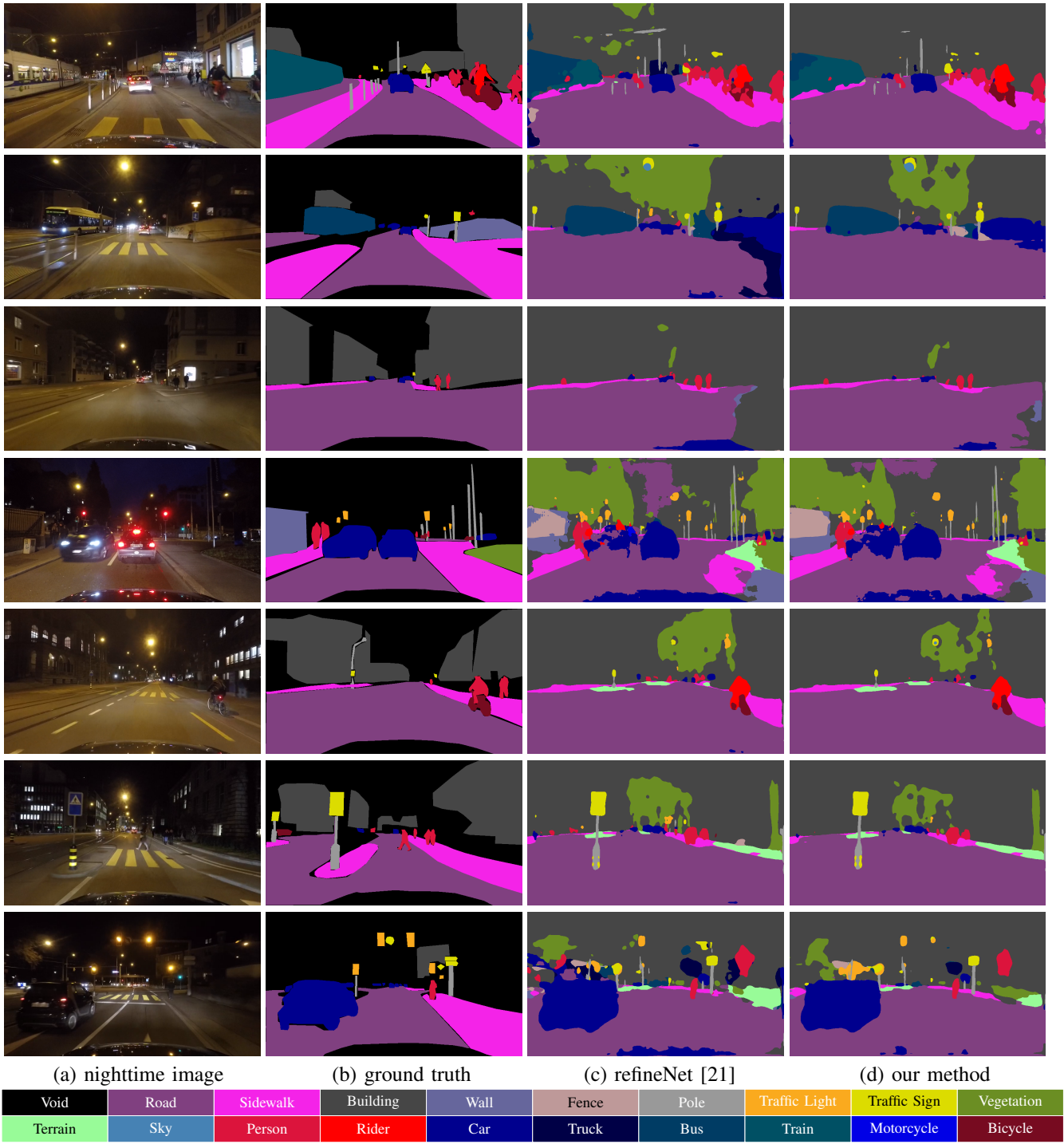
Also, it can be seen from the table that our method benefits from the progressive adaptation in three steps, i.e. from daytime to civil twilight, from civil twilight to nautical twilight, and from nautical twilight to astronomical twilight. The complete pipeline outperforms the two incomplete alternatives. This means that the gradual adaptation closes the domain gap progressively. As the model is adapted one more step forward, the gap to the target domain is further narrowed. Data recorded through twilight time constructs a trajectory between the source domain (daytime) and the target domain (nighttime) and makes daytime-to-nighttime  knowledge transfer feasible.  

Finally, we find that our three-step progressive pipeline outperforms the one-step progressive alternative. This is mainly due to the unsupervised nature of the model adaptation: the method learns from generated labels for model adaptation. This means that the accuracy of the generated labels directly affect the quality of the adaptation. The one-step adaptation alternative proceeds more aggressively and in the end learns from more noisy generated labels than than our three-step complete pipeline. The three-step model adaptation method generate labels only on data which falls slightly off the training domain of the previous model.
Our three-step model adaptation strikes a good balance between computational cost and quality control.  
 
\textbf{Qualitative Results}. We also show multiple segmentation examples by our method (the three-step complete pipeline) and the original daytime RefineNet model in Figure \ref{fig:sem:seg1}. From the two figures, one can see that our method generally yields better results than the original RefineNet model. For instance, in the second image of Figure \ref{fig:sem:seg1}, the original RefineNet model misclassified some \emph{road} area as \emph{car}. 

While improvement has been observed, the  performance of for nighttime scenes is still a lot worse than that for daytime scenes. Nighttime scenes are indeed more challenging than daytime scenes for semantic understanding tasks. There are more underlying causal factors of variation that generated night data, which requires either more
training data or more intelligent learning approaches to disentangle
the increased number of factors. Also, the models are adapted in an unsupervised manner. Introducing a reasonable amount of human annotations of nighttime scenes will for sure improve the results. This constitutes our future work. 

\textbf{Limitation}.
Many regions in nighttime images are uncertain for human annotators. Those areas should be treated as a separate, special class; algorithms need to be trained to predict this special class as well. It is misleading to assign a class label to those areas. This will be implemented in our next work. We also argue that street lamps should be considered as a separate class in addition to the classes considered in Cityscapes' daytime driving.

\section{CONCLUSIONS}
\label{sec:conclusion}
This work has investigated the problem of semantic image segmentation of nighttime scenes from a novel perspective.  This paper has proposed a novel method to \emph{progressive} adapts the semantic models trained on daytime scenes to nighttime scenes via the bridge of twilight time --- the time between dawn and sunrise, or between sunset and dusk. Data recorded during twilight times are further grouped into three subgroups for a three-step progressive model adaptation, which is able to transfer knowledge from daytime to nighttime in an unsupervised manner.  In addition to the method, a new dataset of road driving scenes is compiled. It consists of 35,000 images ranging from daytime to twilight time and to nighttime. Also, 50 diverse nighttime images are densely annotated for method evaluation. The experiments show that our method is effective for knowledge transfer from daytime scenes to nighttime scenes without using human supervision.

\vspace{1mm}
\noindent
\textbf{Acknowledgement} This work is supported by Toyota Motor Europe via the research project TRACE-Zurich. 


%%%%%%%%%%%%%%%%%%%%%%%%%%%%%%%%%%%%%%%%%%%%%%%%%%%%%%%%%%%%%%%%%%%%%%%%%%%%%%%%

%%%%%%%%%%%%%%%%%%%%%%%%%%%%%%%%%%%%%%%%%%%%%%%%%%%%%%%%%%%%%%%%%%%%%%%%%%%%%%%%

%%%%%%%%%%%%%%%%%%%%%%%%%%%%%%%%%%%%%%%%%%%%%%%%%%%%%%%%%%%%%%%%%%%%%%%%%%%%%%%%

%\section*{ACKNOWLEDGMENT}

\bibliographystyle{ieee}
\bibliography{IEEEfull}

%%%%%%%%%%%%%%%%%%%%%%%%%%%%%%%%%%%%%%%%%%%%%%%%%%%%%%%%%%%%%%%%%%%%%%%%%%%%%%%%

\end{document}